\documentclass[sigconf,nonacm]{acmart}

  \hypersetup{%
    pdfauthor={Shuaiwen Li, Weihang Yu, Ke Wang, Meng Han},
    pdftitle={AEC-DS: Adaptive Erasure Coding with PDP-Triggered Reputation and QoS-Aware Migration for Decentralized Storage},
    pdfsubject={arXiv preprint}}

\graphicspath{{./images/}} 

\usepackage{makecell}
\begin{document}

\title{AEC-DS: Adaptive Erasure Coding with PDP-Triggered Reputation and QoS-Aware Migration for Decentralized Storage
}
\author{Shuaiwen Li}
  \orcid{0009-0007-7559-536X}
  \affiliation{%
    \institution{Zhejiang University}
    \city{Hangzhou}
    \country{China}}
  \email{swli@zju-if.com}

  \author{Weihang YU}
  \affiliation{%
    \institution{Central University of Finance and Economics}
    \city{Beijing}
    \country{China}}
  \email{ywh20071028@163.com}

  \author{Ke Wang}
  \orcid{0009-0009-6149-3535}
  \affiliation{%
    \institution{Binjiang Institute of Zhejiang University}
    \city{Hangzhou}
    \country{China}}
  \email{kwang@zju-if.com}

  \author{Meng Han}
  \orcid{0000-0001-7472-0842}
  \affiliation{%
    \institution{Zhejiang University}
    \city{Hangzhou}
    \country{China}}
  \email{mhan@zju.edu.cn}

  \renewcommand{\shortauthors}{Li et al.}



\maketitle

\section{Introduction}
Erasure coding provides durability for decentralized storage, but current schemes still lack feedback between auditing results and redundancy adaptation \cite{li2024sok}. Static-EC cannot reflect node reliability differences; Dynamic-EC \cite{zhang2024dynamic} adjusts redundancy from global failure rates, causing bandwidth overhead during failures; DRD-EC \cite{qiao2025drd} considers reputation in coding decisions but does not integrate audit feedback or proactively move valuable data, resulting in inflexible overhead or delayed recovery.
We present a PDP-driven adaptive mechanism that separates node reputation tiers from data service classes. The feedback loop follows the sequence of audit, reputation update, redundancy recomputation, migration, and re-audit, allowing the system to adjust the parity level and selectively relocate shards as conditions change. Nodes with consistently successful audit results are promoted gradually, whereas nodes with repeated failures receive an additional penalty through the penalty factor.
Experiments on 800 nodes and 500 files achieve $1.25\times$ storage overhead, 66.8\%--75.2\% fewer recoveries, and 100\% durability. Ablation results show that migration contributes 176.8\% to loss prevention, indicating a trade-off between migration bandwidth and reliability.
Our contributions are: (1) a PDP-based closed-loop framework connecting audits with relocation; (2) a QoS-aware dual-track migration strategy for proactive protection; (3) large-scale evaluation and ablation studies analyzing system components.

\section{Related Work}

Our work integrates erasure coding, PDP auditing, and reputation-based placement, which have been studied separately \cite{chen2025fast,zhang2025beyond}. PDP was introduced by Ateniese et al. \cite{ateniese2007pdp} and refined by Shacham and Waters \cite{shacham2008compact}, but audit results were not linked to redundancy decisions.
Existing erasure coding methods \cite{shen2025survey} also lack adaptive feedback. Static-EC \cite{abebe2018ecstore,nicolaou2022ares} uses fixed code rates, causing storage inefficiency and durability risks. Dynamic-EC \cite{zhang2024dynamic} adjusts redundancy using global failure rates, but reacts slowly and ignores data importance. DRD-EC \cite{qiao2025drd} introduces object-level reputation for parity adjustment, yet still lacks proactive shard relocation.
We propose a closed-loop mechanism that combines PDP feedback, reputation modeling, adaptive erasure coding, and QoS-weighted migration. Unlike ``audit $\rightarrow$ detect $\rightarrow$ reconstruct,'' our system enables ``audit $\rightarrow$ predict $\rightarrow$ migrate'' to reduce recovery costs. Table~\ref{tab:differences} summarizes the differences from prior methods \cite{zhang2024dynamic,qiao2025drd}.

\section{System Model and Design}
This section presents the architecture and key algorithms of the proposed closed-loop adaptive erasure coding mechanism. Section~\ref{sec:system_model} introduces the system model and threat assumptions, followed by the node layering architecture and reputation update rules in Section~\ref{sec:reputation}. Section~\ref{sec:control_flow} describes the closed-loop control process from ``audit $\rightarrow$ reputation $\rightarrow$ redundancy adjustment $\rightarrow$ class migration $\rightarrow$ re-audit.'' Section~\ref{sec:migration} presents the promotion and demotion algorithms for shard migration within the dual-track layered architecture, while Section~\ref{sec:authenticity} discusses the co-design approach for off-chain data authenticity assurance.

\begin{table}[t!]

\small
\centering
\caption{Functional comparison with prior work. ($\checkmark$~=~supported, $\times$~=~unsupported)}
\label{tab:differences}
\begin{tabular}{p{5.0cm} c c}
\toprule
\textbf{Capability} & \textbf{Existing Work} & \textbf{Ours} \\
\midrule
Closed-loop feedback
(Audit results directly trigger redundancy \& placement decisions)
& $\times$
& $\checkmark$ \\
\addlinespace
Dual-track decoupling (Node tiers and data classes evolve independently)
& $\times$
& $\checkmark$ \\
\addlinespace
Cross-tier data migration
& $\times$
& $\checkmark$ \\
\addlinespace
Preventive relocation (Triggered by reputation decay, before data loss)
& $\times$
& $\checkmark$ \\
\addlinespace
QoS-aware node selection (Data importance weighted in resource allocation)
& $\times$
& $\checkmark$ \\
\addlinespace
Storage-honesty triggered (PDP audit outcome as the primary decision signal)
& $\times$
& $\checkmark$ \\
\bottomrule
\end{tabular}
\vspace{-15pt}
\end{table}

\subsection{System Model and Threat Assumptions}
\label{sec:system_model}
\textbf{System Model.} We consider a decentralized network of $N$ storage nodes $\mathcal{N} = \{n_1,\dots,n_N\}$ managing $F$ files. Each file is Reed--Solomon \cite{kubiatowicz2000oceanstore,benet2014ipfs,filecoin2017} encoded into $k$ data shards and $m$ parity shards ($n=k+m$), dispersed across distinct nodes. For simplicity we fix $k=4$; the parity count $m \in [m_{\min}, m_{\max}]$ is dynamically determined by our algorithm (typical configuration $m_{\min}=1$, $m_{\max}=4$). Each data object carries a static QoS class $q \in \{\mathrm{H},\mathrm{M},\mathrm{L}\}$ (high, medium, low), specified by the user at creation and mapped to a quantitative priority $\mathrm{QoS} \in (0,1]$.

\textbf{Threat Assumptions.} We consider two types of faulty behavior: (1) benign failures, where nodes may temporarily become unavailable with probability $p_{\text{offline}}$ due to network instability or hardware degradation; (2) Byzantine behavior, where up to a fraction $f_{\text{mal}}$ of nodes may intentionally drop or modify stored data to avoid service costs. Malicious nodes may collude, but cannot generate valid PDP proofs under the adopted cryptographic mechanisms \cite{ateniese2007pdp,shacham2008compact}. We assume the security of homomorphic verifiable tags and random challenges, while the smart contract layer responsible for issuing audit challenges and updating reputation is trusted and operates beyond direct node control.

\subsection{Node Layering and Reputation Model}
\label{sec:reputation}

\textbf{Node Layering Architecture.} Based on dynamic reputation scores $R_i \in [0,1]$, nodes are divided into three logical tiers:

\textbf{Hot tier:} $R_i < 0.7$. Low-reputation or newly joined nodes. They undergo the highest PDP audit frequency (e.g., every 2 hours) for rapid fault detection.

\textbf{Warm tier:} $0.7 \leq R_i \leq 0.95$. Medium-reputation nodes, audited at $1/2$ to $1/3$ of the Hot tier frequency.

\textbf{Cold tier:} $R_i > 0.95$. This tier contains stable nodes with high reputation. These nodes have the lowest audit frequency, which is about one-third of that of the Warm tier, thereby reducing communication overhead.

Node tiers are reviewed periodically, for example, every 24 hours, based on the latest reputation scores. When a node's score crosses a tier threshold, the system updates its tier accordingly. The resulting node pools are then used for subsequent data shard class migration.

\textbf{Reputation Update Model.} A node's reputation score $R_i$ is updated only according to its PDP audit results. When the smart contract challenges node $n_i$, the node must return a proof of possession within the specified timeout. Let $S \in \{0,1\}$ denote the audit result, where $S=1$ represents a successful audit and $S=0$ represents a timeout or an invalid proof. The reputation score is updated using an exponentially weighted moving average:

\begin{equation}
  \label{eq:reputation_update}
  R_i^{(t+1)} = (1-\alpha) R_i^{(t)} + \alpha \cdot S
\end{equation}

where $\alpha \in (0,1)$ is the learning rate that determines the influence of the latest audit result relative to the previous reputation score. We set $\alpha=0.1$ in the experiments. Repeated successful audits gradually increase the reputation score toward $1$. A single failed audit causes only a limited decrease, whereas consecutive failures cause the score to decay over successive updates and may lead to node demotion.

\subsection{Closed-Loop Control Flow}
\label{sec:control_flow}

\begin{figure*}[!htb]
\centering
\includegraphics[width=0.80\linewidth]{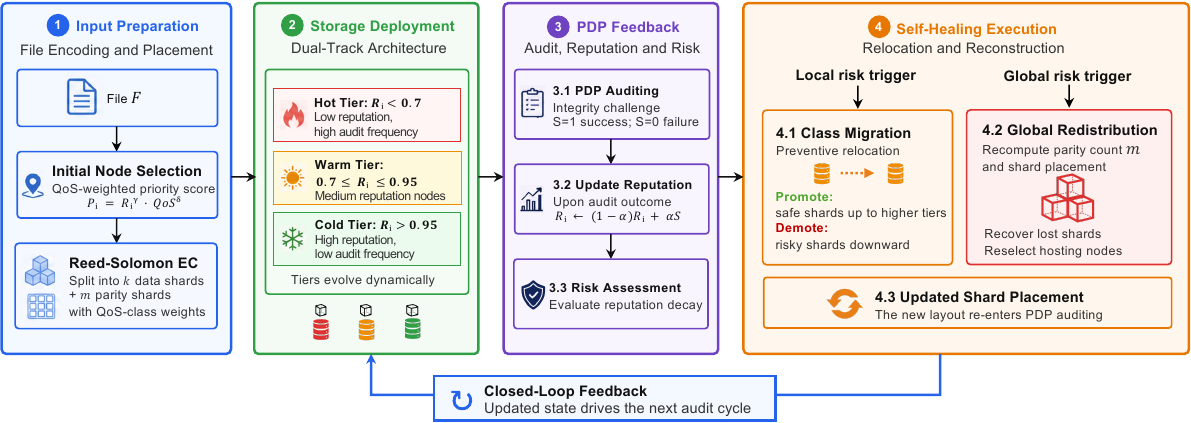}
\caption{Closed-Loop Control with Tier Migration for PDP-Driven Adaptive Erasure Coding}
\label{fig:fig2}
\vspace{-10pt}
\end{figure*}

The mechanism operates as a self-optimizing loop comprising four sequential stages, as illustrated in Figure~\ref{fig:fig2}. This section details the control logic and the adaptive redundancy recomputation, while the execution of data relocation is deferred to Section~\ref{sec:migration}.


\textbf{Stage 1: Differentiated PDP Auditing.} Smart contracts issue integrity challenges at tier-specific intervals. Let the base interval for the Hot tier be \( T_H \). Auditing intervals for the Warm and Cold tiers are scaled by factors \( \lambda_W, \lambda_C > 1 \), such that \( T_{Warm}=\lambda_W T_H \) and \( T_{Cold}=\lambda_C T_H \). This tiered frequency intensively monitors low-reputation nodes while minimizing overhead for stable ones.

\textbf{Stage 2: Real-time Reputation Update.} Once a PDP challenge is resolved, the binary outcome \( S \in \{0,1\} \) (1 for success, 0 for failure or timeout) immediately triggers an on-chain reputation update for the challenged node. The node's reputation score \( R_i \) is recalculated using the exponentially weighted moving average defined in Eq.~\eqref{eq:reputation_update} (Section~\ref{sec:reputation}), ensuring that audit results are reflected in node trust scores.

\textbf{Stage 3: Adaptive Redundancy Re-computation.} The parity count \( m \) for a data object is recalculated to proactively adjust its fault tolerance before actual data loss occurs. Re-computation is triggered when either of two conditions is met:

(i) The mean reputation \( \bar{R} \) of the nodes currently hosting the object's shards drops by more than a threshold \( \theta \) since the last re-computation.

(ii) Any hosting node experiences \( f_{\text{fail}} \) consecutive PDP failures.

The new parity count is computed as:

\begin{equation}
  \label{eq:parity_count}
  m = \lceil m_{\min} + (m_{\max} - m_{\min}) \cdot (1 - \bar{R}) \cdot \delta(\text{QoS}) \rceil \quad
\end{equation}

where

\begin{equation}
  \label{eq:qos_factor}
  \delta(\text{QoS}) = \begin{cases} 1.2, & \text{QoS} > 0.8 \\ 1.0, & 0.4 \le \text{QoS} \le 0.8 \\ 0.8, & \text{QoS} < 0.4 \end{cases} \quad
\end{equation}

\textbf{Stage 4: Shard Relocation and Tier Migration.} Based on the decision made in Stage 3, the system updates the shard placement in one of two ways. A change in \(m\) triggers global redistribution, including shard reconstruction when necessary and reselection of the \(k+m\) hosting nodes. If \(m\) remains unchanged but a node's reputation meets the migration condition, only the affected shards are relocated through incremental class migration. The corresponding procedures are described in Section~3.4.

\textbf{Closed-Loop Nature.} After relocation, the updated shard placement and parity configuration are used in the next PDP auditing cycle. The resulting audit outcomes update node reputations again and may lead to another redundancy adjustment or migration decision. In this way, PDP evidence is incorporated into subsequent coding and placement decisions rather than being used only for failure detection.

\subsection{Redistribution and Class Migration}
\label{sec:migration}

When the parity count \( m \) is updated, the system relocates data to align with the new redundancy level and current node reputations. We distinguish two granularities: global redistribution triggered by \( m \) changes, and incremental migration triggered by reputation fluctuations when \( m \) is stable.

\subsubsection{Global Redistribution} 

When \( m \) changes, all \( n = k + m_{\text{new}} \) hosting nodes are reselected. Lost shards are first recovered via Reed-Solomon decoding using any \( k \) survivors. The \( n \) new nodes are then chosen by a QoS-weighted priority score:

\begin{equation}
  \label{eq:priority}
  P_i = R_i^{\gamma} \cdot \text{QoS}^{\delta} \quad
\end{equation}

where \( \gamma > 1 \) amplifies high-reputation advantage and \( \delta < 1 \) tempers QoS disparity, preventing top-tier monopolization (\( \gamma=1.5, \delta=0.8 \)). The top \( n \) nodes by \( P_i \) host the reconstituted shards.

\subsubsection{Dual-Track Architecture} 

Traditional schemes couple node tiers with data placement, causing two issues: data cannot migrate as trust evolves, and low-reputation nodes lack incentives to improve. We decouple these dimensions:

\textbf{Node-centric track:} Node tiers (Hot, Warm, Cold) evolve solely based on their dynamic reputation scores, independent of the data they store.

\textbf{Data-centric track:} Data objects retain an immutable, user-defined QoS class throughout their lifecycle.

A shard's physical location is thus determined by the dynamic intersection of its data object's static QoS class and its hosting node's current tier. This decoupling enables cross-tier migration on demand without altering the data's inherent class, allowing the system to proactively move high-value shards to more trustworthy nodes as their reputations change.


\subsubsection{Class Migration} 

When \( m \) remains stable but node reputations fluctuate significantly, incremental migration adjusts shard placements preventively using asymmetric strategies:

\textbf{Promotion (Lower to Higher Tier).} A shard on a Hot or Warm node is promoted only if its host passes \( c_{\text{up}} \) consecutive PDP audits and the migration priority \( P_{\text{migrate}} = R_i \cdot \text{QoS} \) exceeds threshold \( \tau_{\text{up}} \) (e.g., 0.88). The shard then moves to the optimal node in the target higher tier via Eq.~\eqref{eq:priority}. This conservative design operates at single-shard granularity and requires sustained honest behavior.

\textbf{Demotion (Higher to Lower Tier).} Demotion rapidly isolates risk through a penalized score:

\begin{equation}
  \label{eq:demotion}
  P_{\text{demote}} = R_i \cdot \text{QoS} \cdot (1 - \lambda \cdot f_{\text{fail}}) \quad
\end{equation}

where \( f_{\text{fail}} \) counts consecutive PDP failures and \( \lambda \in (0.2, 0.3) \) amplifies the penalty. When \( P_{\text{demote}} \) falls below \( \tau_{\text{down}} \) (e.g., 0.65), all shards on that node---prioritized by descending QoS---are immediately evacuated to the highest-scoring nodes in a lower tier.

\subsubsection{Distinction and Synergy} 

Redistribution addresses "is redundancy sufficient?" by global reselection upon \( m \) changes. Migration addresses "where is data safest?" by local adjustment upon reputation shifts. Together they achieve full preventive coverage. Notably, compressed redundancy (\( m \) near \( m_{\min}=1 \)) ensures Cold-tier nodes bear low absolute storage load, while \( \delta < 1 \) in Eq.~\eqref{eq:priority} implicitly balances load across the network by preventing QoS-based monopolization.

\subsection{Off-Chain Data Authenticity Assurance}
\label{sec:authenticity}
AEC-DS uses the verified outcome of each PDP audit as an input to node reputation updates. We adopt a homomorphic-tag-based remote integrity auditing scheme following established PDP/POR constructions \cite{ateniese2007pdp,juels2007pors,shacham2008compact,etemad2016dynamicpdp}. For each audit, the smart contract derives a fresh challenge nonce from on-chain randomness, and the challenged node returns an aggregated proof. The proof is verified without retrieving the original shards, and the fresh nonce prevents a proof prepared for an earlier challenge from being reused. The contract records the binary verification result and updates the reputation of the challenged node accordingly. The security of the underlying proof protocol follows prior work \cite{ateniese2007pdp,li2012cooperative}; this paper focuses on how verified audit outcomes are incorporated into redundancy and placement decisions. In a practical deployment, audit transactions could be batched or submitted through a Layer-2 system to reduce on-chain cost, although these optimizations are not evaluated in our simulations.

\section{Experiments}
We evaluate AEC-DS through event-driven simulations, using storage overhead, cumulative recovery operations, and data durability as the main performance measures. The experiments compare AEC-DS with three baselines, examine the contribution of its individual components, and test its behavior under different adversarial-node ratios. The following subsections present the research questions, simulation settings, baselines, and evaluation metrics.

\subsection{Research Questions}

The evaluation focuses on the following three research questions:

\textbf{RQ1 (Overall Performance):} How does AEC-DS compare with the baseline methods in terms of storage overhead, cumulative recovery operations, and data durability?

\textbf{RQ2 (Closed-Loop Effectiveness):} How much does each component, including class migration, adaptive EC, reputation, PDP feedback, and QoS, contribute to the overall performance? How are recovery operations and storage overhead affected when each component is removed?

\textbf{RQ3 (Robustness):} Does data durability remain stable under different adversarial node ratios? Can the reputation system accurately reflect node reliability?

RQ1 evaluates the overall performance of AEC-DS, RQ2 measures the contribution of each component through ablation experiments, and RQ3 examines the robustness of the system under different conditions.

\subsection{Experimental Setup and Parameters}

\textbf{Simulation Platform.} We developed an event-driven simulator in Python. The simulator models node behavior, PDP audits, reputation updates, RS encoding and decoding, data reconstruction, and shard migration. Each round corresponds to one global audit cycle.

\textbf{Main Configuration.} Table~\ref{tab:sim_params} lists the main parameters used for RQ1 and RQ2.
\begin{table}[!htbp]
  \centering
  \setlength{\abovecaptionskip}{0pt}
  \setlength{\belowcaptionskip}{3pt}

  \caption{Main simulation parameters}
  \label{tab:sim_params}

  \small
  \renewcommand{\arraystretch}{0.95}

  \begin{tabular}{@{}l c@{}}
    \toprule
    \textbf{Parameter} & \textbf{Value} \\
    \midrule
    Nodes $N$                                  & 800 \\
    Files $F$                                  & 500 \\
    Simulation rounds                          & 500 \\
    Independent runs                           & 10 \\
    Adversarial ratio                          & 10\% \\
    Offline probability                        & 0.003/round \\
    Reputation drop threshold $\theta$         & 0.01 \\
    Consecutive failure limit $f_{\text{fail}}$ & 3 \\
    QoS distribution                           & H:35\%, M:45\%, L:20\% \\
    Data shards $k$                            & 4 \\
    Parity range $[m_{\min},m_{\max}]$          & $[1,4]$ \\
    Learning rate $\alpha$                     & 0.1 \\
    Promotion threshold $\tau_{\text{up}}$     & 0.88 \\
    Demotion threshold $\tau_{\text{down}}$    & 0.65 \\
    Penalty factor $\lambda$                    & 0.25 \\
    Priority exponents $(\gamma,\delta)$        & $(1.5,0.8)$ \\
    \bottomrule
  \end{tabular}

  \vspace{-3pt}
\end{table}

\textbf{Supplementary Configuration.} For the robustness analysis in RQ3, we use a smaller and faster configuration, as shown in Table~\ref{tab:fast_main}. This configuration preserves the main experimental trends while reducing the runtime of each simulation by approximately 80\%.

\begin{table}[!htbp]
  \centering
  \setlength{\abovecaptionskip}{0pt}
  \setlength{\belowcaptionskip}{2pt}

  \caption{Fast vs. main configuration}
  \label{tab:fast_main}

  \small
  \begin{tabular}{lcc}
    \toprule
    \textbf{Parameter} & \textbf{Fast Config.} & \textbf{Main Config.} \\
    \midrule
    Simulation rounds & 200 & 500 \\
    Independent runs  & 3   & 10 \\
    Nodes             & 400 & 800 \\
    Files             & 250 & 500 \\
    Node churn        & Off & On \\
    \bottomrule
  \end{tabular}
  \vspace{-5pt}
\end{table}

\subsection{Baselines}
We compare AEC-DS with three baseline methods. Their main differences are summarized in Table~\ref{tab:methods_compare}.

\begin{table*}[h!]
  \caption{Differences between proposed and baseline methods}
  \label{tab:methods_compare}
  \centering
  \small
  \setlength{\tabcolsep}{3pt}
  \resizebox{\textwidth}{!}{%
  \begin{tabular}{l c c c c c}
    \toprule
    \textbf{Method} &
    \textbf{\makecell[c]{Redundancy\\Criterion}} &
    \textbf{Feedback Loop} &
    \textbf{QoS-Aware} &
    \textbf{\makecell[c]{Class\\Migration}} &
    \textbf{Granularity} \\
    \midrule
    Static-EC    & Fixed value          & None                  & No   & No   & --- \\
    Dynamic-EC   & Global failure rate  & Open-loop             & No   & No   & Global \\
    DRD-EC       & Mean reputation      & Unidirectional open-loop & No & No & Per-object \\
    Proposed-EC  & Mean reputation + QoS & Closed-loop           & Yes  & Yes  & Per-object \\
    \bottomrule
  \end{tabular}}
\end{table*}

\textbf{Static-EC} keeps $m=2$ fixed, ignoring node dynamics \cite{abebe2018ecstore,nicolaou2022ares}.

\textbf{Dynamic-EC} sets $m = \lceil m_{\min} + (m_{\max}-m_{\min}) \cdot p_{\text{fail}} \rceil$ per round, but global triggers cause bandwidth storms \cite{zhang2024dynamic} and react to lagging indicators.

\textbf{DRD-EC} computes $m = \lceil m_{\min} + (m_{\max}-m_{\min}) \cdot (1-\bar{R}) \rceil$ using mean reputation, yet lacks a feedback loop and proactive migration \cite{qiao2025drd} , making it our key ablation baseline.

The comparisons isolate gains: Proposed-EC vs. DRD-EC reveals closed-loop and migration benefits; DRD-EC vs. Dynamic-EC \cite{qiao2025drd,zhang2024dynamic} shows reputation-guided over failure-guided adjustment; Dynamic-EC vs. Static-EC captures value of any dynamism.

\subsection{Evaluation Metrics}
We define metrics covering storage efficiency, reliability, and dynamic behavior.  

\textbf{Storage Overhead:} the ratio of total stored data to original size, for RS codes \((k+m)/k\). Closer to 1 indicates higher efficiency. The average across all files is:  

\begin{equation}
  \label{eq:function6}
  \text{StorageOverhead} = \frac{1}{F} \sum_{f=1}^{F} \frac{k + m_f}{k} \quad
\end{equation}

where \(m_f\) is the parity count of file \(f\).  

\textbf{Cumulative Recovery Count:} total reconstruction operations executed during simulation. Unlike prior work reporting instantaneous rates, we use the cumulative count to reflect long-term cost. Lower values indicate more effective prevention.  

\textbf{Data Durability:} the fraction of objects with at least \(k\) surviving shards at any time. An object is lost when missing shards exceed \(m\). We report final durability and monitor its time evolution.

\section{Experimental Results}
This section presents the evaluation results. Section~\ref{sec:performance} reports comprehensive performance comparisons against the three baselines (RQ1); Section~\ref{sec:ablation} quantifies marginal contributions of core components via ablation studies (RQ2); Section~\ref{sec:robustness} analyzes robustness under varying adversarial node ratios (RQ3).

\subsection{Comprehensive Performance Comparison}
\label{sec:performance}
Table~\ref{tab:perf_indicators} reports the means and standard deviations of storage overhead, cumulative recovery count, and final data durability across 10 independent runs. All methods sustain 100\% data durability, indicating that basic redundancy prevents permanent data loss under our fault model. However, storage and recovery costs diverge markedly.

\begin{table}[h!]
  \caption{Key performance indicators (mean $\pm$ std, 10 runs)}
  \label{tab:perf_indicators}
  \centering
  \small
  \setlength{\tabcolsep}{4pt}
  \begin{tabular}{l c c c}
    \toprule
    \textbf{Method} &
    \textbf{\makecell[c]{Storage\\Overhead}} &
    \textbf{\makecell[c]{Cumulative\\Recoveries}} &
    \textbf{Data Durability} \\
    \midrule
    Static-EC   & $1.500 \pm 0.000$ & $4848 \pm 156$   & 1.000 \\
    Dynamic-EC  & $1.760 \pm 0.131$ & $5720 \pm 1410$  & 1.000 \\
    DRD-EC      & $1.250 \pm 0.000$ & $4283 \pm 1599$  & 1.000 \\
    Proposed-EC & $1.250 \pm 0.000$ & $1420 \pm 557$   & 1.000 \\
    \bottomrule
  \end{tabular}
\end{table}

\textbf{Storage Efficiency.} Proposed-EC and DRD-EC achieve the lowest overhead (1.25), reducing storage cost by 16.7\% compared with Static-EC and 29.0\% compared with Dynamic-EC \cite{zhang2024dynamic}. Although both reduce $m$ to $m_{\min}=1$, they employ different adaptation strategies. DRD-EC updates $m$ reactively according to current mean reputation \cite{qiao2025drd}, without preventive migration; when node reliability declines, recovery is required after failures occur. Proposed-EC instead relocates shards to high-reputation Cold-tier nodes in advance. As a result, Proposed-EC requires only 33.2\% of DRD-EC's recoveries (1420 vs. 4283), showing the advantage of proactive migration over reconstruction.

\textbf{Recovery Count.} Proposed-EC's 1420 cumulative recoveries represent reductions of 70.7\% over Static-EC, 75.2\% over Dynamic-EC, and 66.8\% over DRD-EC. Pairwise $t$-tests (Table~\ref{tab:significance}) confirm statistical significance ($p<0.01$) with Cohen's $d$ magnitudes exceeding 2.2 (``very large'' effect sizes), directly evidencing the value of the closed-loop design.

\begin{table}[h!]
  \caption{Statistical significance of recovery count differences (Proposed-EC vs. baselines)}
  \label{tab:significance}
  \centering
  \small
  \setlength{\tabcolsep}{4pt}
  \begin{tabular}{l c c c c}
    \toprule
    \textbf{Comparison} &
    \textbf{Mean Diff.} &
    \textbf{\shortstack{t-statistic}} &
    $p$-value &
    \textbf{\shortstack{Cohen's $d$}} \\
    \midrule
    vs. Static-EC   & $-3428.7$ & $-16.44$ & $<0.001$ & $-7.949$ \\
    vs. Dynamic-EC  & $-4300.7$ & $-6.94$  & $<0.001$ & $-3.805$ \\
    vs. DRD-EC      & $-2863.4$ & $-4.45$  & $0.0016$ & $-2.269$ \\
    \bottomrule
  \end{tabular}
\end{table}

\begin{figure}[!htb]
\centering
\includegraphics[width=0.70\columnwidth]{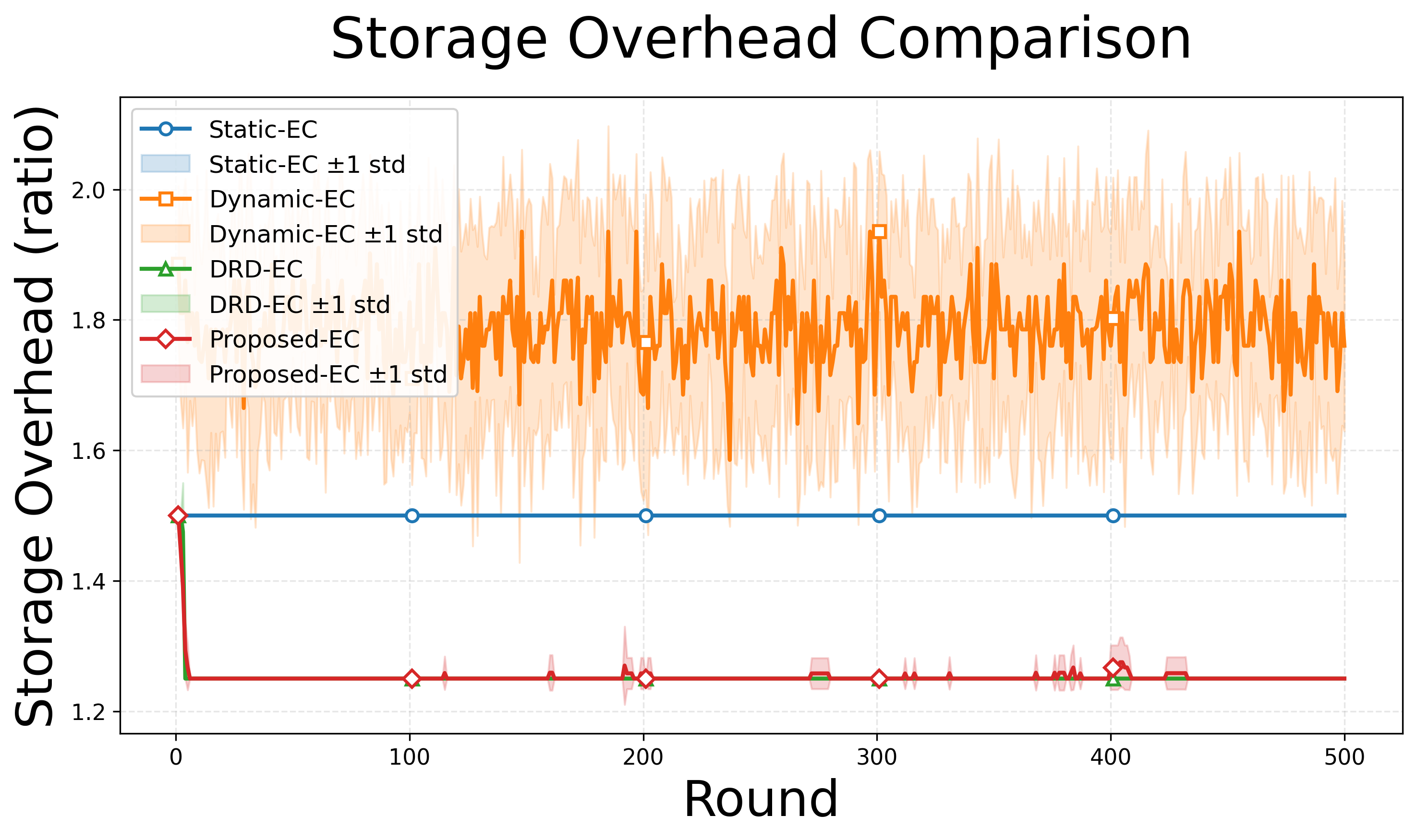}
\caption{Time Evolution of Storage Overhead for Various Methods}
\label{fig:fig3}
\vspace{-5pt}
\end{figure}

\begin{figure}[!h]
\centering
\includegraphics[width=0.70\columnwidth]{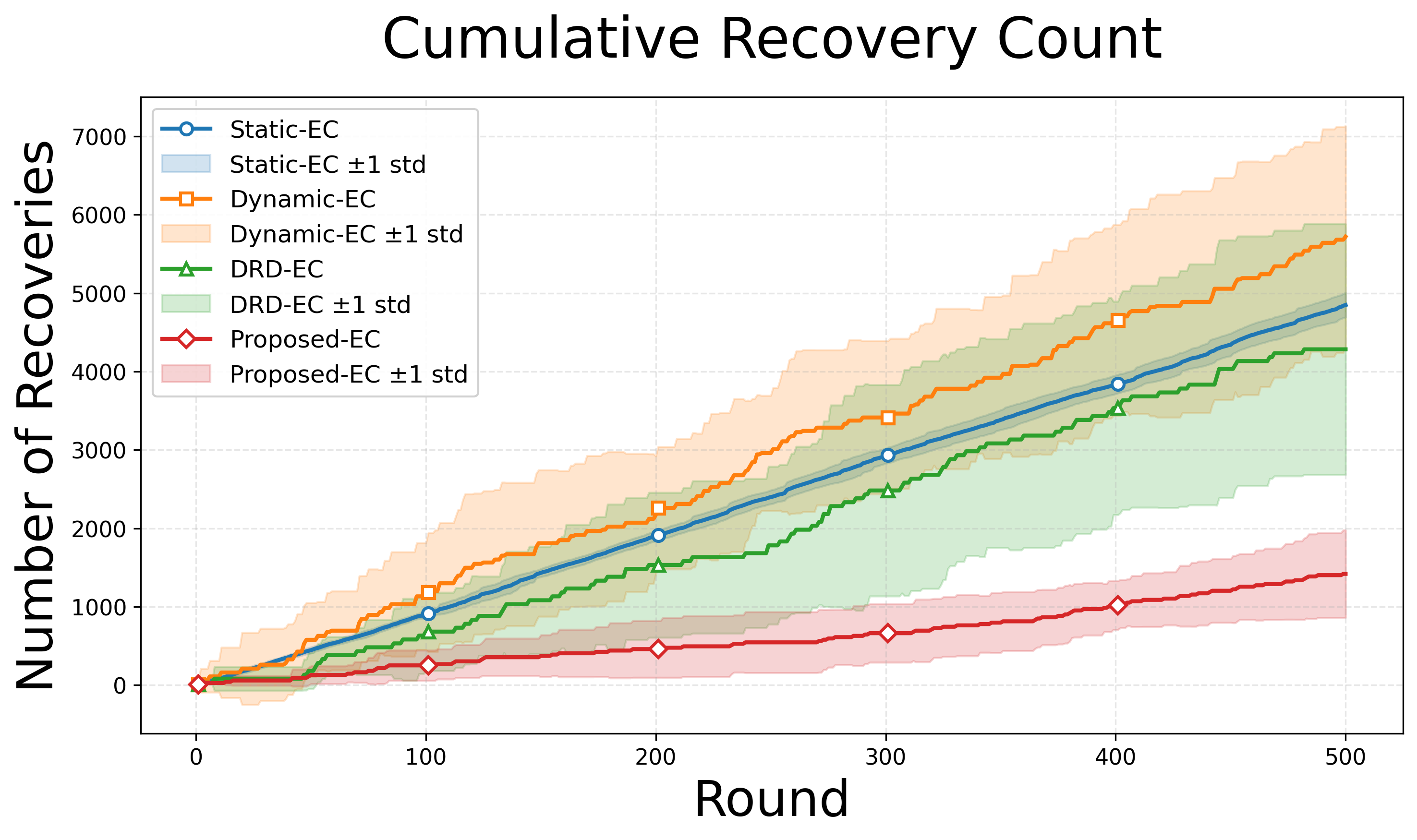}
\vspace{-5pt}
\caption{Cumulative Recovery Counts for Various Methods}
\label{fig:fig4}
\vspace{-10pt}
\end{figure}

\textbf{Time Evolution.} Storage overhead converges from 1.5 to 1.25 within 30--50 rounds for both Proposed-EC and DRD-EC (Figure~\ref{fig:fig3}). The cumulative recovery curves (Figure~\ref{fig:fig4}) show a clear difference: DRD-EC recoveries increase almost linearly, while Proposed-EC exhibits a slower growth trend. This difference reflects the impact of closed-loop adaptation. Without continuous audit feedback, DRD-EC accumulates recovery operations as failures increase; in contrast, Proposed-EC updates placement decisions over time and reduces future recovery demand.

\subsection{Ablation Study}
\label{sec:ablation}
\begin{figure}[!h]
\centering
\includegraphics[width=0.70\columnwidth]{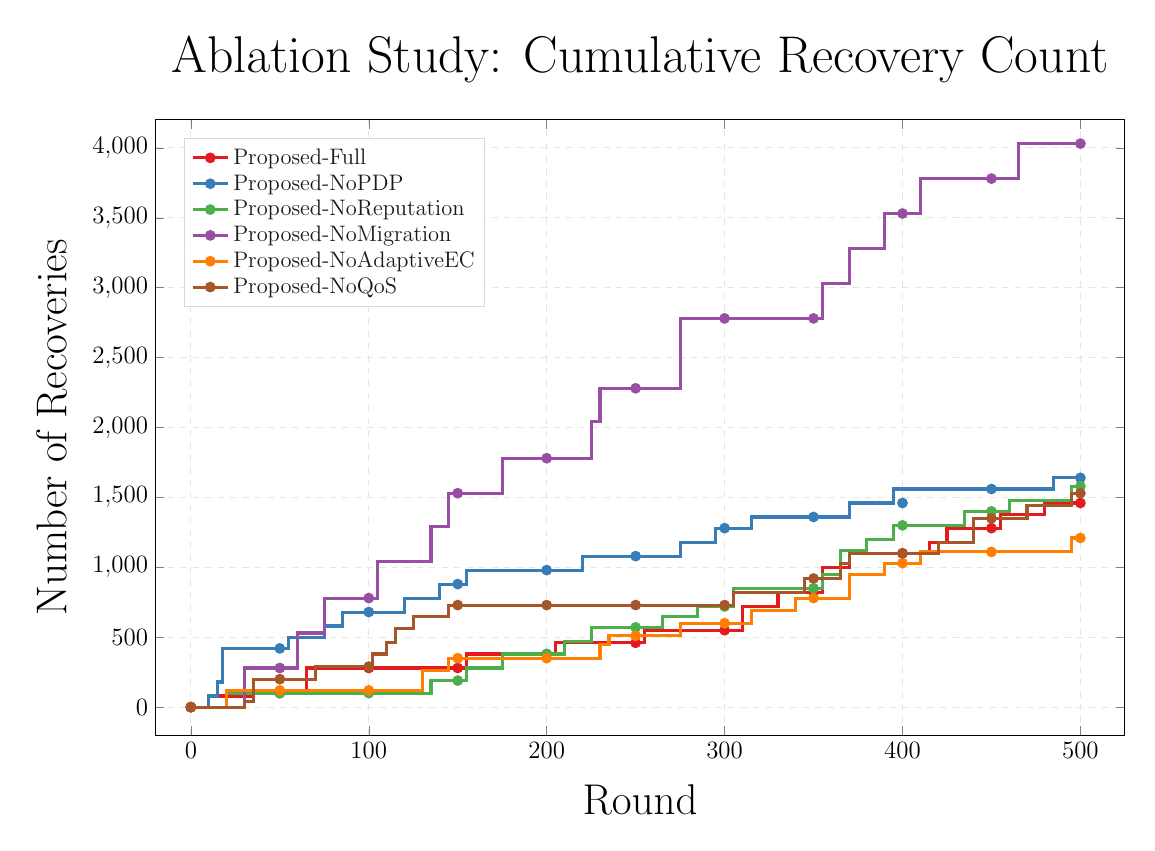}
\vspace{-5pt}
\caption{Comparison of Recovery Counts}
\label{fig:fig5}
\vspace{-10pt}
\end{figure}

We quantify the marginal contribution of each closed-loop component by sequentially removing PDP feedback (NoPDP), reputation (NoReputation), class migration (NoMigration), adaptive EC (NoAdaptiveEC), and QoS differentiation (NoQoS). Table~\ref{tab:impact_ranking} ranks components by their comprehensive impact; Figure~\ref{fig:fig5} plots corresponding cumulative recovery curves.

\begin{table}[h!]
  \centering
  \small
  \setlength{\abovecaptionskip}{0pt}
  \setlength{\belowcaptionskip}{3pt}
  \caption{Component impact ranking}
  \label{tab:impact_ranking}
  \centering
  \small
  \setlength{\tabcolsep}{9pt}
  \begin{tabular}{c l c }
    \toprule
    \textbf{Rank} & 
    \textbf{\makecell[c]{Removed  Component}} & 
    \textbf{\makecell[c]{Recovery Count Change}} \\
    \midrule
    1 & Class Migration      & $+176.8\%$  \\
    2 & Adaptive EC          & $-17.0\%$   \\
    3 & Reputation System    & $+8.5\%$    \\
    4 & PDP Feedback         & $+12.7\%$   \\
    5 & QoS Differentiation  & $+5.0\%$    \\
    \bottomrule
  \end{tabular}
  \vspace{-15pt}
\end{table}

\textbf{Class migration} is the dominant pillar: removing it (NoMigration) inflates recoveries from 1,457.5 to 4,035.0 (+176.8\%), the severest degradation. This quantifies how proactively evacuating at-risk shards drastically reduces passive reconstruction.

\begin{figure}[!h]
\centering
\includegraphics[width=0.70\columnwidth]{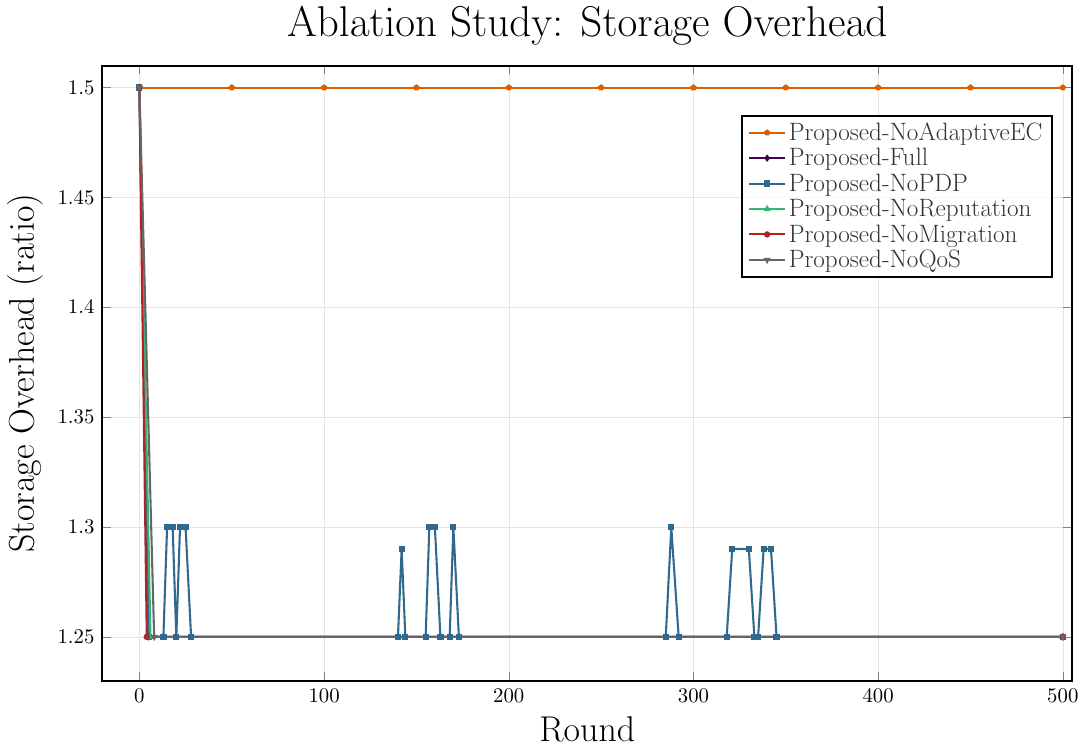}
\vspace{-10pt}
\caption{Comparison of Storage Overhead}
\label{fig:fig6}
\vspace{-10pt}
\end{figure}

\textbf{Adaptive EC} determines storage efficiency. Figure~\ref{fig:fig6} compares the storage overhead among all variants. Removing adaptive EC (NoAdaptiveEC) fixes ($m=2$), increasing overhead from 1.25 to 1.50 (+20\%), while reducing recoveries by 17.0\% due to higher redundancy. The results illustrate the storage--recovery trade-off: adaptive EC improves storage utilization while maintaining reliability, although it introduces additional algorithmic complexity.

\textbf{Reputation} is used to distinguish nodes with different reliability levels during placement. In the NoReputation variant, the storage overhead rises to 1.50 and the number of recoveries increases by 8.5\%. Without reputation scores, the system cannot distinguish node reliability when making placement and redundancy decisions.

\textbf{PDP feedback} provides recent audit results for system adjustment. Removing this feedback (NoPDP) increases the number of recoveries by 12.7\%. This suggests that timely audit results help the system identify risky nodes and take preventive action. At the same time, the auditing process may introduce additional communication overhead.

\textbf{QoS differentiation} has the smallest effect, with a 5.0\% increase in recoveries after it is removed. Its main role is to set different placement priorities for data objects, so that high-value data can be assigned to more reliable nodes. Therefore, its effect on the overall metrics is relatively limited.

Overall, class migration and adaptive EC determine where shards are placed and how much redundancy is used. Reputation and PDP feedback provide information for these decisions, while QoS differentiation adjusts the priority of different data objects.

\subsection{Robustness Analysis}
\label{sec:robustness}
We evaluate the system with adversarial-node ratios of 5\% and 10\% using the fast configuration, and each result is averaged over three runs. As shown in Table~\ref{tab:robustness_adv}, the system maintains 100\% data durability in both settings. When the adversarial ratio increases from 5\% to 10\%, the final mean reputation decreases from 0.938 to 0.911, while cumulative recoveries increase from 1043.6 to 1222.9. These results show that a higher adversarial ratio leads to lower reputation scores and greater recovery demand, while no permanent data loss is observed under the tested settings.

\begin{table}[h!]
  \centering
  \small
  \setlength{\abovecaptionskip}{0pt}
  \setlength{\belowcaptionskip}{3pt}
  \caption{Robustness to adversarial ratio (fast config., 3-run means)}
  \label{tab:robustness_adv}
  \centering
  \small
  \setlength{\tabcolsep}{4pt}
  \begin{tabular}{c c c c c}
    \toprule
    \textbf{\makecell[c]{Adversarial \\ Ratio}} &
    \textbf{\makecell[c]{Storage \\ Overhead}} &
    \textbf{\makecell[c]{Final Mean\\Reputation}} &
    \textbf{\makecell[c]{Cumulative\\Recovers}} &
    \textbf{\makecell[c]{Final \\ Durability}} \\
    \midrule
    5\%  & 1.25 & 0.938 & 1043.6 & 1.000 \\
    10\% & 1.25 & 0.911 & 1222.9 & 1.000 \\
    \bottomrule
  \end{tabular}
  \vspace{-10pt}
\end{table}

\section{Discussion}

\subsection{Trade-off and Applicable Scenarios}
AEC-DS uses additional migration traffic to reduce recovery demand, lowering cumulative recoveries by 66.8\%--75.2\%. It is therefore more suitable for bandwidth-tolerant decentralized storage, such as edge, Web3, and P2P networks \cite{filecoin2017,benet2014ipfs,merlec2024blockchainstorage,reno2025storjledger}, but less suitable for bandwidth-limited or latency-sensitive environments.

\subsection{Limitations}
The current evaluation is simulation-based, uses a fixed \(k=4\), omits economic incentives, and relies on manually selected parameters. Future work will include testbed validation, automatic parameter tuning, and staking mechanisms. Overall, AEC-DS combines PDP feedback, adaptive erasure coding, and dual-track migration, achieving a \(1.25\times\) storage overhead and 66.8\%--75.2\% fewer recoveries while maintaining 100\% durability. These results suggest that additional migration traffic can improve storage reliability when sufficient bandwidth is available.

\section*{Acknowledgments}
  This research was supported by National Key R\&D Program of China (Grant No. 2023YFB2704400).

\bibliographystyle{ACM-Reference-Format}

\bibliography{references}

@inproceedings{ateniese2007pdp,
  author    = {Ateniese, Giuseppe and Burns, Randal and Curtmola, Reza and Herring, Joseph and Kissner, Lea and Peterson, Zachary and Song, Dawn},
  title     = {Provable Data Possession at Untrusted Stores},
  booktitle = {Proceedings of the 14th ACM Conference on Computer and Communications Security},
  year      = {2007},
  pages     = {598--609},
  doi       = {10.1145/1315245.1315318}
}

@inproceedings{juels2007pors,
  author    = {Juels, Ari and Kaliski, Burton S.},
  title     = {{PORs}: Proofs of Retrievability for Large Files},
  booktitle = {Proceedings of the 14th ACM Conference on Computer and Communications Security},
  year      = {2007}
}

@inproceedings{shacham2008compact,
  author    = {Shacham, Hovav and Waters, Brent},
  title     = {Compact Proofs of Retrievability},
  booktitle = {Advances in Cryptology -- ASIACRYPT 2008},
  year      = {2008},
  publisher = {Springer}
}

@misc{benet2014ipfs,
  author       = {Benet, Juan},
  title        = {{IPFS} - Content Addressed, Versioned, {P2P} File System},
  year         = {2014},
  howpublished = {arXiv:1407.3561},
  url          = {https://arxiv.org/abs/1407.3561}
}

@misc{filecoin2017,
  author       = {{Protocol Labs}},
  title        = {Filecoin: A Decentralized Storage Network},
  year         = {2017},
  url          = {https://research.protocol.ai/publications/filecoin-a-decentralized-storage-network/protocollabs2017a.pdf}
}

@inproceedings{kubiatowicz2000oceanstore,
  author    = {Kubiatowicz, John and others},
  title     = {OceanStore: An Architecture for Global-Scale Persistent Storage},
  booktitle = {Proceedings of the Ninth International Conference on Architectural Support for Programming Languages and Operating Systems},
  year      = {2000}
}

@article{li2024sok,
  author  = {Li, C. and others},
  title   = {{SoK}: Decentralized Storage Network},
  journal = {Blockchain: Research and Applications},
  year    = {2024},
  url     = {https://www.sciencedirect.com/science/article/pii/S2667295224000424}
}

@article{shen2025survey,
  author  = {Shen, Z. and Cheng, K. and others},
  title   = {A Survey of the Past, Present, and Future of Erasure Coding for Storage Systems},
  journal = {ACM Transactions on Storage},
  year    = {2025}
}

@inproceedings{abebe2018ecstore,
  author    = {Abebe, M. and others},
  title     = {{EC-Store}: Bridging the Gap between Storage and Latency in Distributed Erasure Coded Systems},
  booktitle = {Proceedings of the IEEE International Conference on Distributed Computing Systems},
  year      = {2018}
}

@article{nicolaou2022ares,
  author  = {Nicolaou, N. and others},
  title   = {{ARES}: Adaptive, Reconfigurable, Erasure coded, Atomic Storage},
  journal = {ACM Transactions on Storage},
  year    = {2022}
}

@article{zhang2024dynamic,
  author  = {Zhang, Ming and Wu, Chen and Li, Jia and Guo, Minyi},
  title   = {Dynamic-EC: An efficient dynamic erasure coding method for permissioned blockchain systems},
  journal = {Frontiers of Computer Science},
  year    = {2024},
  doi     = {10.1007/s11704-023-3209-3}
}

@inproceedings{qiao2025drd,
  author    = {Qiao, Jiehan and Zhang, Jinnan and Chen, Rurong and others},
  title     = {{DRD-EC}: a dynamic risk-driven erasure coding method for {IoT} blockchains},
  booktitle = {Proceedings of SPIE, Volume 13692: International Conference on Computer Application and Information Security (ICCAIS 2024)},
  year      = {2025},
  volume    = {13692},
  doi       = {10.1117/12.3069027}
}

@misc{etemad2016dynamicpdp,
  author       = {Etemad, M. and others},
  title        = {A Generic Dynamic Provable Data Possession Framework},
  year         = {2016},
  howpublished = {Cryptology ePrint Archive, Paper 2016/748},
  url          = {https://eprint.iacr.org/2016/748}
}

@article{li2012cooperative,
  author  = {Zhu, Yan and Hu, Hongxin and Ahn, Gail-Joon and Yu, Mengyang},
  title   = {Cooperative Provable Data Possession for Integrity Verification in Multicloud Storage},
  journal = {IEEE Transactions on Parallel and Distributed Systems},
  year    = {2012}
}

@article{merlec2024blockchainstorage,
  author  = {Merlec, M.M. and others},
  title   = {Blockchain-Based Decentralized Storage Systems for Sustainable Data Self-Sovereignty: A Comparative Study},
  journal = {Sustainability},
  year    = {2024}
}

@article{reno2025storjledger,
  author  = {Reno, S. and others},
  title   = {StorjLedger: An Innovative Distributed Storage Ecosystem That Overcomes Blockchain Trade-Offs Using Erasure-Coded Sharding and Proof-of-Storage Consensus},
  journal = {Engineering Reports},
  year    = {2025}
}

@article{chen2025fast,
  title={A Fast Consensus Algorithm of Large-Scale Heterogeneous Dynamic IoT Nodes for DAG-based Blockchain},
  author={Chen, Yourong and Zhuang, Yubo and Wang, Shiwei and Guo, Yidan and Han, Meng and Liu, Liyuan and Hong, Zhen},
  journal={IEEE Internet of Things Journal},
  year={2025},
  publisher={IEEE}
}

@article{zhang2025beyond,
  title={Beyond contrastive learning: adaptive graph representations with mutual information maximization for blockchain and structured data},
  author={Zhang, Yifeng and Ren, Qianqian and Chen, Yourong and Han, Meng},
  journal={Complex \& Intelligent Systems},
  volume={11},
  number={9},
  pages={412},
  year={2025},
  publisher={Springer}
}

\end{document}